\PassOptionsToPackage{table}{xcolor}
\documentclass[letterpaper, 10 pt, journal, twoside]{IEEEtran}

\IEEEoverridecommandlockouts                              

\usepackage{graphics} 
\usepackage{epsfig} 
\usepackage{mathptmx} 
\usepackage{amsmath} 
\usepackage{amssymb}  
\usepackage{bm}
\usepackage{cite}

\usepackage{booktabs}
\usepackage{caption}
\usepackage{etoolbox}
\usepackage{graphicx}
\usepackage{multicol}
\usepackage{multirow}
\usepackage{pifont}
\usepackage{xcolor}
\definecolor{gray}{RGB}{229, 230, 230}
\usepackage{hyperref}
\usepackage{bbm}
\usepackage{subcaption}

\hypersetup{
	colorlinks=true,
	linkcolor=green,
	filecolor=magenta,      
	urlcolor=magenta,
	citecolor=green,
}

\newcommand{\PAR}[1]{\vskip4pt \noindent{\bf #1~}}

\begin{document}
	\title{MonoTher-Depth: Enhancing Thermal Depth Estimation via Confidence-Aware Distillation}
	\author{Xingxing Zuo, Nikhil Ranganathan, Connor Lee, Georgia Gkioxari, and Soon-Jo Chung
    \thanks{Manuscript received: 9 September, 2024; Revised 6 December, 2024; Accepted 6 January, 2025. This paper was recommended for publication by Editor Cesar Cadena upon evaluation of the Associate Editor and Reviewers' comments. This work was supported by the Office of Naval Research. (\textit{Corresponding author: Xingxing Zuo.})}
	\thanks{All authors are with California Institute of Technology, Pasadena, California, USA. E-mail:{\tt\footnotesize \{zuox, nrangana, clee, georgia, sjchung\}@caltech.edu}}%
    \thanks{Digital Object Identifier (DOI): see top of this page.}
	}
	

	\maketitle
	\begin{abstract}
Monocular depth estimation (MDE) from thermal images is a crucial technology for robotic systems operating in challenging conditions such as fog, smoke, and low light. The limited availability of labeled thermal data constrains the generalization capabilities of thermal MDE models compared to foundational RGB MDE models, which benefit from datasets of millions of images across diverse scenarios. To address this challenge, we introduce a novel pipeline that enhances thermal MDE through knowledge distillation from a versatile RGB MDE model. Our approach features a confidence-aware distillation method that utilizes the predicted confidence of the RGB MDE to selectively strengthen the thermal MDE model, capitalizing on the strengths of the RGB model while mitigating its weaknesses. Our method significantly improves the accuracy of the thermal MDE, independent of the availability of labeled depth supervision, and greatly expands its applicability to new scenarios.
In our experiments on new scenarios without labeled depth, the proposed confidence-aware distillation method reduces the absolute relative error of thermal MDE by 22.88\% compared to the baseline without distillation. The code will be available at: \url{https://github.com/ZuoJiaxing/monother_depth}.
\end{abstract} 
	\vspace{-12pt}\section{Introduction}
\label{sec:introduction}

\begin{figure*}[t]
  \centering
  \includegraphics[width=0.9\textwidth]{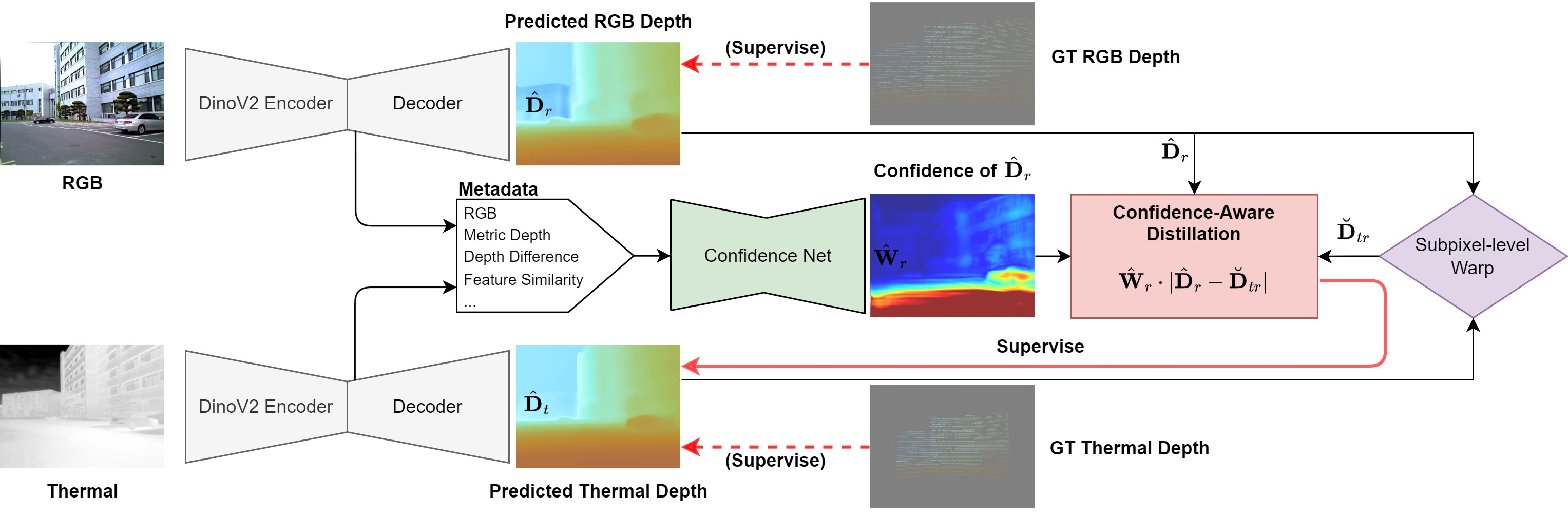}
  \caption{\textbf{System architecture of MonoTher-Depth.} Our framework enhances the thermal MDE model by leveraging learned priors from an RGB model. To harness the strengths and mitigate the weaknesses of the RGB teacher model, we predict the confidence of its depth output $\hat{\mathbf{W}}_r$ using curated metadata that includes both thermal and RGB information. Whether ground-truth (GT) depth is available or not, our system improves thermal MDE through confidence-aware distillation by minimizing the confidence-weighted depth discrepancy between the predicted RGB depth $\hat{\mathbf{D}}_r$ and the wrapped thermal depth $\breve{\mathbf{D}}_{tr}$.}
  \vspace{-1em}
  \label{fig:system-design}
\end{figure*}
%


%

\IEEEPARstart{D}{epth} estimation is a fundamental problem in various applications, including autonomous driving, robotics, and mixed reality. Monocular depth estimation (MDE) from a single RGB camera is widely used and has seen significant progress recently~\cite{rajapaksha2024deep}.
Existing RGB MDE methods have achieved high accuracy, detailed fidelity, and great zero-shot generalization capabilities.
A key factor contributing to the success of RGB MDE models is the abundance of existing RGB datasets with labeled depth available for training.
For instance, methods such as UniDepth~\cite{piccinelli2024unidepth}, Metric3D~\cite{yin2023metric3d}, and ZeroDepth~\cite{guizilini2023towards} are trained on datasets containing 3M, 8M, and 17M images with labeled depth, respectively. DepthAnything~\cite{yang2024depth} leverages 62M unlabeled images, on top of 1.5M labeled images, to achieve great generalization in diverse environments.

However, RGB cameras struggle in adverse visual conditions characterized by low light, fog, or smoke, which limit the performance of MDE in these scenarios. Thermal cameras, which capture long-wave infrared signals, can penetrate atmospheric particles and provide reliable measurements in obscured and low-light conditions. Despite these advantages, thermal depth estimation has yet to be thoroughly explored. Specifically, thermal MDE inherits the challenges of RGB MDE, with added complexity due to typically low contrast, high signal-to-noise ratio, and a lack of texture and color information in thermal imagery.

Along with these challenges, thermal MDE must contend with the scarcity of labeled thermal datasets. This is in contrast to the abundance of datasets, both real and synthetic, for RGB MDE. In this work, we seek to enhance thermal MDE by taking advantage of off-the-shelf RGB MDE foundation models that have been pretrained on massive RGB datasets. In particular, we propose a novel framework that distills an RGB MDE model to thermal using a confidence-aware approach (Fig.~\ref{fig:system-design}).
Unlike existing RGB-T works~\cite{shivakumar2019pst900,xu2020u2fusion,deevi2024rgb}, our approach can work with RGB-T training image pairs that are not perfectly co-registered. We achieve this by adaptively guiding the distillation process using confidence derived from cross-modal features and depth consistency.

Our main contributions are summarized as follows:
\begin{itemize}
	\item We introduce MonoTher-Depth, a novel semi-supervised distillation framework that distills an RGB MDE model to create a thermal MDE model.
	\item We propose confidence-aware distillation, based on cross-modal spatial consistency of feature spaces and depth estimates, to limit incorrect guidance and to eschew the need for co-registered RGB-T image pairs. 
	\item We perform extensive validation and ablation studies on the MS$^2$~\cite{shin2023deep} and ViViD++~\cite{lee2022vivid++} datasets, demonstrating MonoTher-Depth's effectiveness in enhancing thermal MDE learning with and without ground-truth depth supervision. In scenarios without labeled depth, our proposed confidence-aware distillation method reduces the absolute relative error of thermal MDE by 22.88\% compared to the baseline without distillation. 
\end{itemize}

	\vspace{-12pt}\section{Related Work}
\label{sec:relatedwork}
\PAR{RGB MDE.}
Monocular depth estimation (MDE) from RGB images has made significant strides in recent years~\cite{eigen2014depth, fu2018deep, ranftl2020towards, zoedepth2023, liu2023va, yin2023metric3d, hu2024metric3d, ke2024repurposing,  shao2024iebins, piccinelli2024unidepth}, demonstrating zero-shot capabilities across diverse image datasets. MiDaS~\cite{ranftl2020towards} is a pioneering work that leverages a large collection of diverse datasets for training relative MDE model, showcasing a certain degree of zero-shot capability. 
Zoedepth~\cite{zoedepth2023} focuses on metric MDE with a dedicated metric bins module, pre-trained on 12 datasets using relative depth and fine-tuned on two datasets using metric depth.
MetricDepth~\cite{yin2023metric3d, hu2024metric3d} addresses the metric ambiguity in MDE by transforming all training data into a canonical camera space with a fixed focal length.
UniDepth~\cite{piccinelli2024unidepth} introduces a pseudo-spherical output space representation that effectively disentangles the camera parameters from the depth estimation process.
Marigold~\cite{ke2024repurposing} harnesses the rich priors captured in recent generative diffusion models to achieve generalizable and accurate MDE.
DepthAnything~\cite{yang2024depth} pretrains the MDE model with labeled data and then utilizes large-scale unlabeled data to learn robust representations, enhancing zero-shot capability and overall robustness.

\PAR{Thermal MDE.}
Thermal MDE is significantly less explored in the existing literature, with only a limited number of datasets available for thermal MDE. Most existing thermal MDE methods are self-supervised~\cite{shin2022maximizing, shin2023self}, relying on reconstructing thermal image sequences by predicted depth and poses between images. 
Some approaches also leverage the knowledge from RGB MDE models or fuse RGB information to enhance thermal MDE performance.
Xu \emph{et al.}~\cite{xu2024unveiling} proposed fusing the predictions of an RGB MDE model and a thermal MDE model to obtain refined depth estimation in both day and night scenarios.
Shin \emph{et al.}~\cite{shin2021self} and Guo \emph{et al.}~\cite{guo2023unsupervised} proposed unsupervised multi-spectrum stereo methods for thermal MDE, which supervise the thermal MDE model using self-reconstruction consistency loss calculated from the predicted depth and poses of sequential thermal and/or RGB images in videos.
Shin \emph{et al.}~\cite{shin2023self} adapted an RGB MDE model to the thermal domain in a self-supervised manner. They used impaired RGB and thermal videos to regress image poses and depth, reconstructing each image sequence within its respective domain, while enforcing that the RGB and thermal encoders produce indistinguishable feature maps through an adversarial loss.
In contrast, our proposed confidence-aware distillation method does not require video sequences for training. Instead, it operates directly on the predicted depth, eliminating the need for complex discriminators operating on high-dimensional feature spaces.

\PAR{Knowledge Distillation for MDE.} 
Knowledge distillation was initially proposed for image recognition tasks~\cite{hinton2015distilling}, but a few methods have also utilized it for MDE. 
Several studies~\cite{guo2018learning, tosi2019learning} have explored supervising an MDE model using predictions from a stereo matching network. 
Pilzer \emph{et al.}~\cite{pilzer2019refine} proposed enhancing the MDE model by refining the predicted depth based on the cycle inconsistency of left-right stereo image pairs.
Aleotti \emph{et al.}~\cite{aleotti2020real} introduced a method to distill a complex MDE model trained on a large-scale dataset into a lightweight model suitable for deployment on handheld devices. 
Poggi \emph{et al.}~\cite{poggi2020uncertainty} proposed a self-teaching strategy to quantify the uncertainty in self-supervised MDE trained from sequential video frames.
URCDC-Depth~\cite{shao2023urcdc} cross-distills Transformer and CNN-based MDE networks by feeding the same image into both networks and fusing their depth predictions based on predicted uncertainty.
Shi \emph{et al.}~\cite{shi20233d} fused MDE results from multiple video frames to build a 3D mesh via TSDF-Fusion, using rendered depth from the mesh to fine-tune the MDE network.
DepthAnything~\cite{yang2024depth} trained a teacher MDE model with labeled depth and then distilled the trained model into a student model using a mix of labeled and massive unlabeled data. During distillation, strongly perturbed images are fed into the student model, while the unperturbed image is fed into the teacher model, allowing the student model to learn robustness to open-world images.
In contrast to the methods mentioned above, our MonoTher-Depth focuses on cross-modality distillation from RGB to thermal. More importantly, our distillation process is confidence-aware, selectively absorbing the strengths of the teacher model while mitigating its weaknesses, and does not require sequential video frames for training.


	\vspace{-5pt}\section{Methodology}

\subsection{Problem Setup}
To overcome the challenges caused by a lack of co-registered RGB-T training datasets for thermal MDE, we distill a pretrained RGB MDE model into a thermal MDE model. Our training method requires RGB-T images with overlapping field-of-views and calibrated extrinsics, but does not need strict co-registration. The RGB model is not needed for inference.


While the RGB depth model performs well on average, its performance may vary across different image regions and conditions. To avoid transferring potentially incorrect behaviors to the thermal model, we model the confidence of RGB predictions and use it to adaptively steer the thermal model during training. This confidence is generated by a dedicated neural network, which is trained when ground-truth depth is available. When ground-truth depth is not available, our method leverages the frozen RGB model and confidence network to modulate the distillation loss of the thermal MDE model. 


\vspace{-10pt}\subsection{Metric MDE Network}
We adopt the metric version of DepthAnything as both our RGB and thermal MDE models~\cite{yang2024depth}. This model uses DinoV2~\cite{dinov2} for feature extraction, a DPT decoder~\cite{ranftl2020towards} for relative depth prediction, and a metric bins module~\cite{zoedepth2023} to produce metric depth predictions. For our RGB teacher model, we use the pretrained DepthAnything model, which was trained on 63.5M images, showing exceptional performance across multiple MDE benchmarks.  

The thermal MDE model utilizes the same architecture, but incorporates an additional preprocessing step that normalizes each raw 16-bit thermal image to the 2\textsuperscript{nd} and 98\textsuperscript{th} percentiles~\cite{lee2024caltech, 10342016}. While some learning-based thermal works~\cite{shin2022maximizing, lee2024caltech, reza2004realization, 10342016} also use Contrast Limited Adaptive Histogram Equalization (CLAHE), we did not find benefits for our MDE purpose.  


\vspace{-10pt}\subsection{Sub-pixel Warp of Thermal-RGB Depth}
In order for the thermal MDE model to learn from the RGB MDE model, a precise spatial mapping between thermal and RGB pixels is essential. We compute this using predicted depth maps and known camera intrinsics and extrinsics. Given the predicted depth from the RGB MDE model, $\hat{\mathbf{D}}_r$, we transform it into the thermal image plane as follows:
\begin{align}
\hat{\mathbf{D}}_{rt} , \hat{\mathbf{u}}_{rt}  = \pi( \mathbf{T}^t_r \hat{\mathbf{D}}_r \pi^{-1}( \mathbf{u}_r, \mathbf{K}_r), \mathbf{K}_t)
\label{eq:transform}
\end{align}
where $\pi(\mathbf{x}, \mathbf{K})$ denotes the camera projection function that projects points $\mathbf{x}$ into the image plane with camera intrinsic matrix $\mathbf{K}$. The inverse projection function $\pi^{-1}(\mathbf{u}, \mathbf{K})$ maps image pixels $\mathbf{u}$ back into the 3D unit plane.   The transformation matrix $\mathbf{T}^t_r$ represents the 6-DoF extrinsic transformation from the RGB camera to the thermal camera. In this equation, we omit the homogeneous conversion of vectors for simplicity. 
Using the transformation function in~\eqref{eq:transform}, we obtain the pixel correspondences $\mathbf{u}_r$ in the RGB image and $\hat{\mathbf{u}}_{rt}$ in the thermal image. The depth $\hat{\mathbf{D}}_{rt}$ represents the RGB depth in the thermal image's coordinate frame.

Similarly, we can transform the predicted thermal depth $\hat{\mathbf{D}}_t$ into the RGB image plane using:
\begin{align}
	\hat{\mathbf{D}}_{tr} , \mathbf{u}_{tr}  = \pi( \mathbf{T}^r_t \hat{\mathbf{D}}_t \pi^{-1}( \mathbf{u}_t, \mathbf{K}_t), \mathbf{K}_r)
	\label{eq:transform1}
\end{align}



To supervise the thermal MDE using the RGB model, we calculate the warped thermal depth corresponding to the RGB image, denoted as $\breve{\mathbf{D}}_{tr}$.
To achieve sub-pixel accuracy in the depth warp process, we sample the depth values of $\hat{\mathbf{D}}_{tr}$ at the sub-pixel locations $\hat{\mathbf{u}}_{tr}$ using bilinear interpolation:
\begin{align}
	\breve{\mathbf{D}}_{tr} = f{_{\rm{bilinear}}}(\hat{\mathbf{u}}_{rt}, \hat{\mathbf{D}}_{tr})
\end{align}


\vspace{-10pt}\begin{figure}[ht]
	\centering
	\includegraphics[width=\columnwidth]{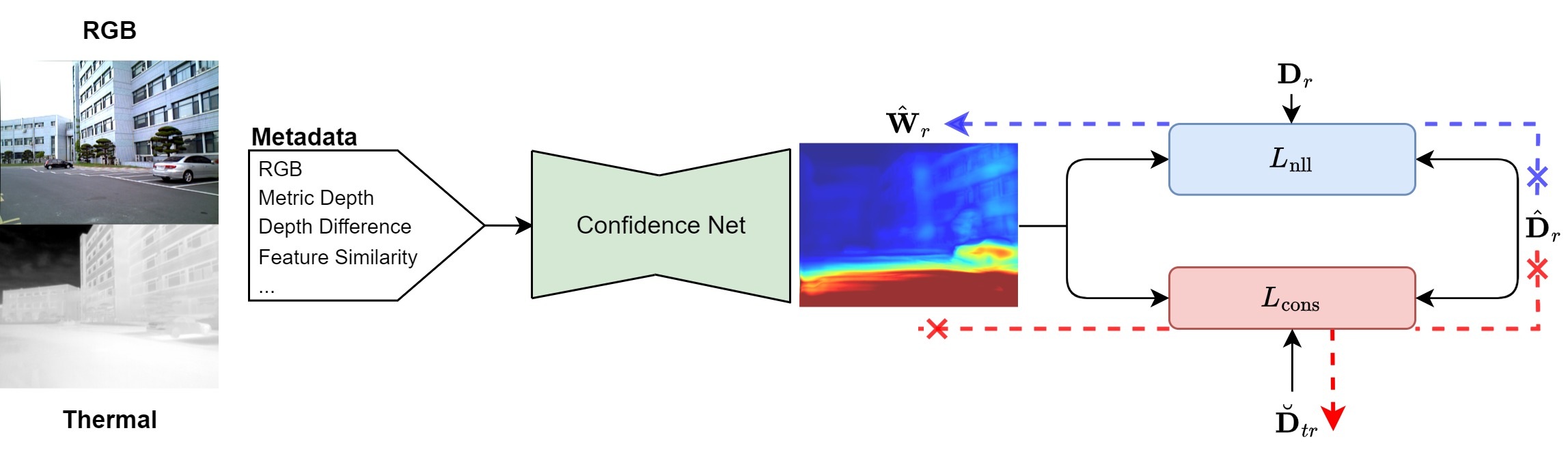}
       \vspace{-1em}
	\caption{\textbf{Pipeline of the confidence-aware distillation.} The predicted confidence $\hat{\mathbf{W}}_r$ of the RGB depth $\hat{\mathbf{D}}_r$ plays a key role in both the negative log-likelihood loss $L_{\rm{nll}}$~(\ref{eq:nll}) and the consistency loss  $L_{\rm{con}}$ (\ref{eq:con}). The $L_{\text{nll}}$ loss propagates gradients back to the confidence network, while the $L_{\text{con}}$ loss propagates gradients to the warped thermal depth $\breve{\mathbf{D}}_{tr}$. Gradient flow is stopped along all other paths.}
    \vspace{-1em}
	\label{fig:conf_training}
\end{figure}

\vspace{-10pt}\subsection{Confidence-Aware Model Distillation} 

We leverage the versatile RGB MDE model to instruct the thermal MDE model at the depth output level by minimizing the discrepancy between the predicted RGB depth $\hat{\mathbf{D}}_r$ and the warped thermal depth $\breve{\mathbf{D}}_{tr}$. Since the reliability of RGB depth predictions may vary across different image regions, weighting all pixel-wise depth discrepancies equally during training can lead to poor results. To address this, we incorporate the confidence of the RGB MDE model into the distillation process.

We design a U-Net to predict the confidence $\hat{\mathbf{W}}_r$ of the RGB MDE output $\hat{\mathbf{D}}r$. To achieve this, we input RGB-aligned metadata into this network, leveraging the pixel correspondences $\mathbf{u}_r$ and $\hat{\mathbf{u}}_{rt}$ between the RGB and thermal depths. This metadata includes: (I) the cosine distance between the RGB feature map and its corresponding thermal feature map, (II) the sampled cosine distance between the thermal feature map and its corresponding RGB feature map, (III) the ${L}_1$ distance $ | \hat{\mathbf{D}}_r - \breve{\mathbf{D}}_{tr}|$, (IV) the warped thermal depth $\breve{\mathbf{D}}_{tr}$, (V) the RGB depth $\hat{\mathbf{D}}_r$, and (VI) the RGB image $\mathbf{I}_r$. Notably, all these curated metadata components are subpixel-aligned with the RGB image to facilitate precise confidence map prediction.

It is straightforward to obtain metadata components (III)–(VI). To calculate component (I), we extract the RGB and thermal feature embeddings from the last layer of the metric bins module, denoted as $\mathbf{F}_r$ and $\mathbf{F}_t$, respectively. The cosine distance between these feature maps is computed as follows:
\begin{align}
	\mathbf{S}_r = \langle \mathbf{F}_r, f_{\rm{bilinear}}(\hat{\mathbf{u}}_{rt}, \mathbf{F}_t) \rangle
	\label{eq:s_r}
\end{align}
where $\langle \mathbf{A}, \mathbf{B} \rangle$ represents the element-wise cosine distance between feature maps $\mathbf{A}$ and $\mathbf{B}$. In this context, $f_{\rm{bilinear}}(\mathbf{u}, \mathbf{F})$ refers to the bilinear interpolation sampling of $\mathbf{F}$ at pixel locations $\mathbf{u}$ unless otherwise specified.
Similarly, we calculate the cosine distance $\mathbf{S}_t$. Since $\mathbf{S}_t$ is aligned with the thermal image, we perform bilinear interpolation again to obtain component (II), denoted as $\mathbf{S}_{tr} = f_{\rm{bilinear}}(\hat{\mathbf{u}}_{rt}, \mathbf{S}_t)$.

The confidence network is based on a standard U-Net architecture, specifically tailored for confidence estimation. The encoder consists of four downsampling layers, each reducing the spatial dimensions by half while increasing the number of feature channels. The decoder mirrors the encoder's structure, employing four upsampling layers that progressively integrate higher-level features with corresponding lower-level features from the encoder via skip connections. This process gradually restores the spatial resolution of the input metadata. The confidence $\hat{\mathbf{W}}_r$ is finally predicted by a convolutional layer followed by a Sigmoid activation function.


Training of this confidence predictor is done only when ground-truth RGB depth ${\mathbf{D}}_r$ is available (Fig.~\ref{fig:conf_training}). 
To do this, we minimize the negative log-likelihood of a Laplacian distribution:
\begin{align}
  	L_{\rm{nll}} = \frac{1}{N}\sum_{i} \hat{W}^i_r \cdot | \rm{sg}(\hat{{D}}^i_r) - {{D}}^i_r | - \beta \rm{log}(\hat{W}^i_r)
  	\label{eq:nll}
\end{align}
where $\rm{sg}(\cdot)$ denotes the stop-gradient operation to prevent backpropagation through the ground truth. The ground-truth RGB depth is often sparse in outdoor scenarios, and $i$ indexes all the $N$ pixel locations where ground-truth RGB depth is available.


To perform confidence-aware distillation, we minimize the confidence-weighted $L_1$ loss:
\begin{align}
	L_{\rm{cons}} = \frac{1}{M} \sum_{j} \rm{sg}(\hat{W}^j_r)  \cdot | \rm{sg}(\hat{{D}}^j_r) - \breve{{D}}^j_{tr}|
	\label{eq:con}
\end{align} 
where $j$ indexes all $M$ pixel locations $\breve{D}^j_{tr}$ that fall within the RGB image after warping. To address issues due to occlusion (where the warped thermal depth may not have corresponding pixels in the RGB depth map) and imperfect ground-truth depth, we exclude the top 20\% residuals in both~\eqref{eq:nll} and~\eqref{eq:con}. Lastly, we apply a mask for the consistency loss \eqref{eq:con}, considering only the pixel locations with the top 80\% of feature similarity $\mathbf{S}_r$~\eqref{eq:s_r}.


\vspace{-10pt}\subsection{Implementation Details}
\label{sec:impl_details}
Our distillation framework is versatile: it allows for replacement of the DepthAnything model, while maintaining compatibility with the confidence network that processes MDE model outputs.
%
%
When ground-truth depth is available, we supervise the thermal and RGB MDE models using the SILOG loss~\cite{eigen2014depth, bhat2021adabins, zoedepth2023}
\begin{align}
	L_{\rm{silog}} = \sum_{i} \sqrt{\frac{1}{N} \sum_i (g^i)^2-\frac{\lambda}{N^2}\left(\sum_i g^i\right)^2}
\end{align}
Here, $g^i = \log \hat{D}^i - \log D^i$, and $N$  is the number of pixels with valid ground-truth depth in the image. We set $\lambda=0.15$ in our experiments. 


In addition, we align depth and image discontinuities by regularizing the predicted depth $\hat{\mathbf{D}}$ with a smoothness loss~\cite{godard2019digging}
\begin{align}
	 L_{\rm{sm}}(\hat{\mathbf{D}}) = \nabla_x \hat{\mathbf{D}} \cdot \exp \left(-\nabla_x \mathbf{I}_r \right) +  \nabla_y \hat{\mathbf{D}} \cdot \exp \left(-\nabla_y \mathbf{I}_r \right)
\end{align}
We only regularize the predicted RGB depth because applying this loss to thermal depth is potentially detrimental.


When ground-truth depth is available, we train the RGB and thermal MDE models, along with the confidence network, using the following combined loss:
\begin{multline}
	L = L_{\rm{silog}\_r} + L_{\rm{silog}\_t} + \alpha \cdot L_{\rm{cons}} \\ + \beta \cdot L_{\rm{nll}}
            + \gamma \cdot L_{\rm{sm}}(\hat{\mathbf{D}}_r) + \lambda \cdot L_{\rm{sm}}(\hat{\mathbf{W}}_r)
\end{multline}
where $L_{\rm{sm}}(\hat{\mathbf{W}}_r)$ denotes the smoothness loss on the predicted confidence map. We set $\alpha=0.2, \beta=0.1, \gamma=0.01, \lambda=0.001$.

When ground-truth depth is unavailable, we freeze the weights of a pretrained RGB MDE model and confidence network. We use the predicted RGB depth and confidence to train the thermal MDE model, applying only the depth consistency loss $L_{\rm{cons}}$ in this self-supervised fine-tuning process.

	\section{Experiments}

\begin{figure*}[t]
  \centering
  \includegraphics[width=\textwidth]{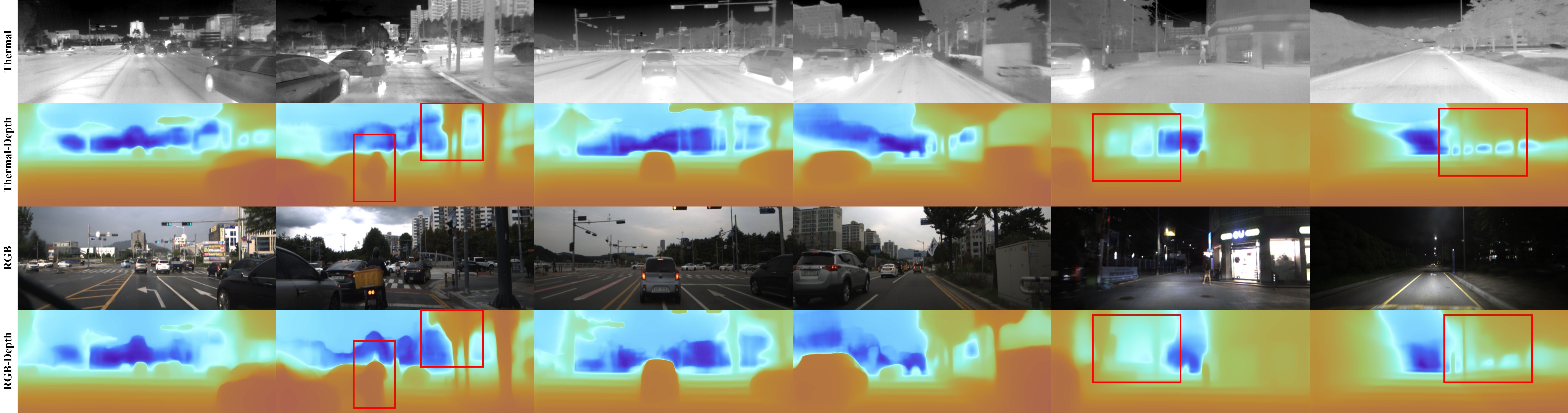}
  \caption{\textbf{Monocular Depth Estimation on MS$^2$ dataset}~\cite{shin2023deep}. Top to bottom: normalized thermal image, predicted thermal depth, RGB image, and predicted RGB depth. Red boxes highlight significant differences between the thermal and RGB depth predictions. Left to right: every two columns showcase the rainy, day, and night conditions, respectively.}
  \vspace{-1em}
  \label{fig:ms2}
\end{figure*}

\subsection{Datasets}\label{sec:experiments}
\PAR{MS$^2$:} This is a multispectral stereo dataset~\cite{shin2023deep} that provides synchronized thermal and RGB images, and projected LiDAR depth maps for benchmarking depth prediction. For MDE benchmarking, we use the left RGB images, left thermal images, and the LiDAR maps. We follow the official train/val/test splits. The train split consists of 7.6K image pairs, while the test split includes 2.3K, 2.3K, and 2.5K image pairs under \textit{day}, \textit{night}, and \textit{rainy} conditions, respectively.
Due to misalignment between the projected ground-truth LiDAR depth and the image—particularly at image edges—training directly with the provided LiDAR depth maps often produces blurred depth predictions. To mitigate this issue, we filter the LiDAR depth map using two strategies: (i) remove depths that exhibit significant pixel-intensity inconsistencies when back-projecting LiDAR points into both the left and right images, and (ii) remove depths that substantially deviate from those obtained via stereo matching~\cite{lipson2021raft}. While we utilize these filtered LiDAR depths during training to achieve sharper predicted depths, we continue to use the unfiltered, officially provided LiDAR depth for evaluation to ensure fairness and thoroughness. Although this setting yields crisper depth predictions, we find it slightly compromises certain evaluation metrics.

\PAR{ViViD++:} This dataset~\cite{lee2022vivid++} provides data from RGB, thermal, and event cameras, as well as LiDAR. It captures both indoor and outdoor scenes. For our study, we only use the official outdoor splits. The outdoor training split, which consists of data collected during the day, is optionally used for self-supervised fine-tuning of our method (see Sec.~\ref{sec:impl_details}). The test split consists of nighttime data to show the generalization of our method.



\vspace{-10pt}\subsection{Training Details}

We initialized the RGB and thermal MDE network encoders with pretrained weights from DepthAnything~\cite{yang2024depth}. We randomly initialized the decoder, the metric bins module of the MDE models, and the confidence network. We used the AdamW optimizer with a learning rate of 8.5e-5 and a weight decay of 0.01. We applied brightness and contrast jitters for data augmentation. 
The networks are trained for 5 epochs on two Nvidia RTX6000 Ada GPUs, with an input image resolution of $256 \times 640$ on MS$^2$ dataset, taking approximately 20 hours. For experiments on the ViViD++ dataset, the input image resolution is $480 \times 640$.


\vspace{-10pt}\subsection{Evaluation Protocols}

We report standard metrics~\cite{zoedepth2023, shin2023deep}, including absolute relative error (AbsRel), squared relative difference (SqRel), root mean square error (RMSE, in meter), RMSE logarithm (RMSElog), and the threshold accuracy $\delta = \%$ of pixels s.t. $\rm{max}(d_i/\hat{d}_i, \hat{d}_i/d_i ) < 1.25^n, n = \{1, 2, 3\}$. 

In outdoor scenarios with ground-truth depth from accumulated LiDAR scans, most pixels with valid ground-truth depth are close to the camera. Simple averaging of metrics over all valid pixels can skew results, as performance on nearby points may dominate. To address this, we report both unweighted metrics averaged over all valid pixels and weighted metrics~\cite{vankadari2024dusk}, which calculate averages across depth bins. Each bin spans 5 meters, and the evaluation covers depths from 0 to 80 meters for both the MS$^2$ and ViViD++ datasets.



\begin{table*}[t!]
	\begin{center}
 \caption{\protect\label{tab:result_ms2} \protect\textbf{Quantitative evaluation of MDE results with various methods on the MS$^2$ dataset~\protect\cite{shin2023deep}.} (Unweighted metrics) 
		}
		\renewcommand\arraystretch{0.95}
		\setlength\tabcolsep{8pt} 
		\small
		\begin{tabular}{c|c|cccc|ccc}
			\hline
			\multirow{2}{*}{\textbf{Methods}}  &\multirow{2}{*}{\textbf{Modality}} &
			\multicolumn{4}{c|}{Error $\downarrow$}&
			\multicolumn{3}{c}{Accuracy $\uparrow$}\\
			
			\cline{3-9}
			&&AbsRel&SqRel&RMSE&RMSElog&$\delta<1.25$&$\delta<1.25^{2}$&$\delta<1.25^{3}$\\
			\hline
			\hline
			DORN~\cite{fu2018deep} & Ther & 0.109 & 0.540 & 3.66 & 0.144 & 0.887 & 0.982 & 0.997 \\
			BTS~\cite{lee2019big} & Ther & 0.086 & 0.380 & 3.163 & 0.117 &  0.926& 0.990 & 0.998 \\
			Adabins~\cite{bhat2021adabins}& Ther & 0.088 & 0.377 & 3.152 & 0.119 & 0.924 & 0.990 & 0.998 \\
			NeWCRF~\cite{yuan2022new} & Ther & 0.080 & 0.331 & 2.937 & 0.109& 0.937 & 0.993 & 0.999 \\
			ZoeDepth~\cite{zoedepth2023} & Ther & 0.091 & 0.425 & 3.202 &  0.123 & 0.915 & 0.989 & 0.998 \\
			DepthAnything~\cite{yang2024depth} &Ther &  0.075&  0.287 & 2.719 &  0.103 & 0.945 & \textbf{0.995} & \textbf{0.999} \\
			Ours & Ther & \textbf{0.072} & \textbf{0.275} &  \textbf{2.677} &\textbf{0.100} & \textbf{0.949} & \textbf{0.995} & \textbf{0.999} \\
			\hline
		\end{tabular}

	\end{center}
\end{table*}

\vspace{-12pt}\begin{table*}[thb!]
	\begin{center}
 \caption{\protect\label{tab:different_weather}\protect\textbf{Detailed quantitative evaluations under `Day', `Night' and `Rainy' conditions on the MS$^2$ dataset~\protect\cite{shin2023deep}.} We show both the unweighted (UnW.) and weighted (W.) metrics.
		}
		\renewcommand\arraystretch{0.99}
		\setlength\tabcolsep{8pt} 
		\small
		\begin{tabular}{c|c|c|cccc|ccc}
			\hline
			\multirow{2}{*}{\textbf{Metrics}} &
			\multirow{2}{*}{\textbf{TestSet}}  &
			\multirow{2}{*}{\textbf{Method}}&
			\multicolumn{4}{c|}{Error $\downarrow$}&
			\multicolumn{3}{c}{Accuracy $\uparrow$}\\
			
			\cline{4-10}
			&&&AbsREL&SqRel&RMSE&RMSElog&$\delta<1.25$&$\delta<1.25^{2}$&$\delta<1.25^{3}$\\
			\hline
			\hline
			\multirow{9}{*}{UnW.} &\multirow{3}{*}{Day}
			&DepthAnything~\cite{yang2024depth} & 0.063 & 0.222 & 2.477 & 0.091 & 0.959 & 0.996 & \textbf{0.999}\\
			&&Zoedepth~\cite{zoedepth2023} & 0.078 & 0.342 & 2.979 & 0.110 & 0.932 & 0.992 & \textbf{0.999}  \\
			&&Ours & \textbf{0.059} & \textbf{0.210} & \textbf{2.420} & \textbf{0.087} & \textbf{0.965} &  \textbf{0.997} & \textbf{0.999} \\
			\cline{2-10}
			
			&\multirow{3}{*}{Night} 
			&DepthAnything~\cite{yang2024depth} & 0.077 & 0.276 & 2.565 & 0.102 & 0.944 & \textbf{0.996} & \textbf{1.000}\\
			&&Zoedepth~\cite{zoedepth2023} & 0.087 & 0.344 & 2.827 & 0.114 & 0.927 & 0.993 & 0.999 \\
			&& Ours & \textbf{0.075} & \textbf{0.268} & \textbf{2.541} & \textbf{0.101} & \textbf{0.945} & \textbf{0.996} & \textbf{1.000}\\
			\cline{2-10}
			&\multirow{3}{*}{Rainy}
			&DepthAnything~\cite{yang2024depth} & 0.085 & 0.358 & 3.085 & 0.115 & 0.932 & 0.992 & \textbf{0.999} \\
			&&Zoedepth~\cite{zoedepth2023} & 0.107 & 0.577 & 3.754 & 0.143 & 0.888 & 0.983 & 0.996\\
			&& Ours & \textbf{0.080} & \textbf{0.342} & \textbf{3.041} & \textbf{0.111} & \textbf{0.939} & \textbf{0.993} & \textbf{0.999} \\
			\hline
			\hline
			\multirow{9}{*}{W.}&\multirow{3}{*}{Day}
			&DepthAnything~\cite{yang2024depth} & 0.083 & 0.553 & 4.181 & 0.104 & 0.935 & 0.995 & 0.999 \\
			&&Zoedepth~\cite{zoedepth2023} & 0.100 & 0.804 & 5.015 & 0.124 & 0.894 & 0.989 & 0.999 \\
			&& Ours & \textbf{0.079} & \textbf{0.538} & \textbf{4.126} & \textbf{0.099} & \textbf{0.942} & \textbf{0.996} & \textbf{1.000} \\
			\cline{2-10}
			
			&\multirow{3}{*}{Night}
			&DepthAnything~\cite{yang2024depth} & 0.091 & 0.603 & 4.081 & 0.107 & 0.924 & 0.994 & \textbf{1.000} \\
			&&Zoedepth~\cite{zoedepth2023} &  0.100 & 0.713 & 4.483 & 0.117 & 0.907 & 0.992 & 0.999\\
			&& Ours &  \textbf{0.088} & \textbf{0.597} & \textbf{4.075} & \textbf{0.105} & \textbf{0.925} & \textbf{0.995} & \textbf{1.000}  \\
			\cline{2-10}
			&\multirow{3}{*}{Rainy} 
			&DepthAnything~\cite{yang2024depth} & 0.100 & 0.724 & 4.768 & 0.122 & 0.905 & 0.991 & \textbf{0.999} \\
			&&Zoedepth~\cite{zoedepth2023} &  0.122 & 1.055 & 5.750 & 0.148 & 0.844 & 0.983 & 0.997  \\
			&& Ours & \textbf{0.096} & \textbf{0.715} & \textbf{4.755} & \textbf{0.118} & \textbf{0.910 }&  \textbf{0.992} & \textbf{0.999} \\
			\hline
		\end{tabular}
		\vspace{-1em}
	\end{center}
\end{table*}

\begin{figure}[tb]
  \centering
  \includegraphics[width=\columnwidth]{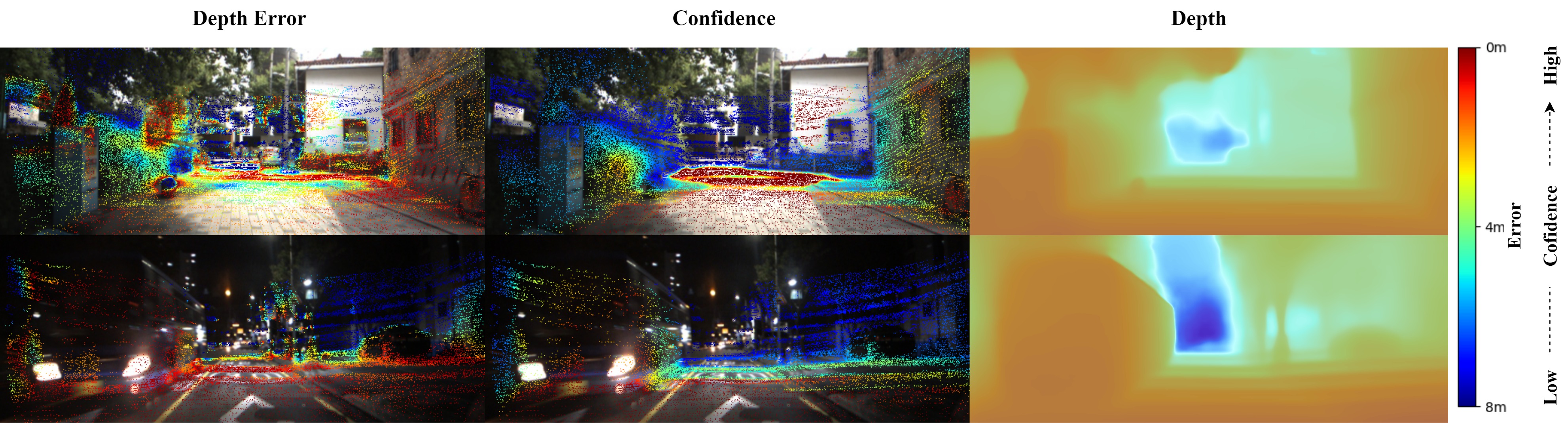}
  \caption{\textbf{Predicted confidence and depth error on MS$^2$ dataset~\cite{shin2023deep}.} Left to right: depth error overlaid on the image, confidence overlaid on the image, and predicted RGB depth.}
  \vspace{-1em}
  \label{fig:viconf}
\end{figure}

\subsection{Evaluation of Metric Monocular Depth Estimation}
\PAR{Evaluation Results on MS$^2$.} Table~\ref{tab:result_ms2} presents evaluation results for metric monocular depth estimation on the MS$^2$ dataset test split. In all tables throughout this paper, we use bold to highlight the best performance. All compared methods were trained or fine-tuned on thermal data of the MS$^2$ training split. In comparison to methods trained exclusively on thermal images~\cite{fu2018deep, lee2019big, bhat2021adabins, yuan2022new, zoedepth2023, yang2024depth}, our MonoTherDepth model performs the best across all metrics.



Our method outperforms both Zoedepth~\cite{zoedepth2023} and DepthAnything~\cite{yang2024depth}, which are closely related to our approach and trained with the thermal images and ground-truth thermal depth supervision.
The improvement of our method is attributed to the effective knowledge distillation from the RGB MDE model during training. While the ground-truth thermal and RGB depth maps from LiDAR offer highly accurate depth information and potentially reduce the benefit of the RGB teacher, the results still highlight the substantial benefits of our confidence-aware distillation. Additionally, Table~\ref{tab:different_weather} provides additional categorized evaluations under \textit{Day}, \textit{Night}, and \textit{Rainy} conditions. Both unweighted and weighted metrics are presented.
Figure~\ref{fig:ms2} presents the MDE results from our method, highlighting the advantages of thermal MDE over RGB MDE in challenging scenarios.
To give an impression of the predicted confidence of the RGB depth, we visualize the depth error and the confidence in Fig.~\ref{fig:viconf}. Interestingly, the regions with large depth errors (blue) coincide with the low-confidence areas (blue), indicating that the confidence map can inclusively reflect the depth error.



\begin{figure*}[t]
  \centering
  \includegraphics[width=\textwidth]{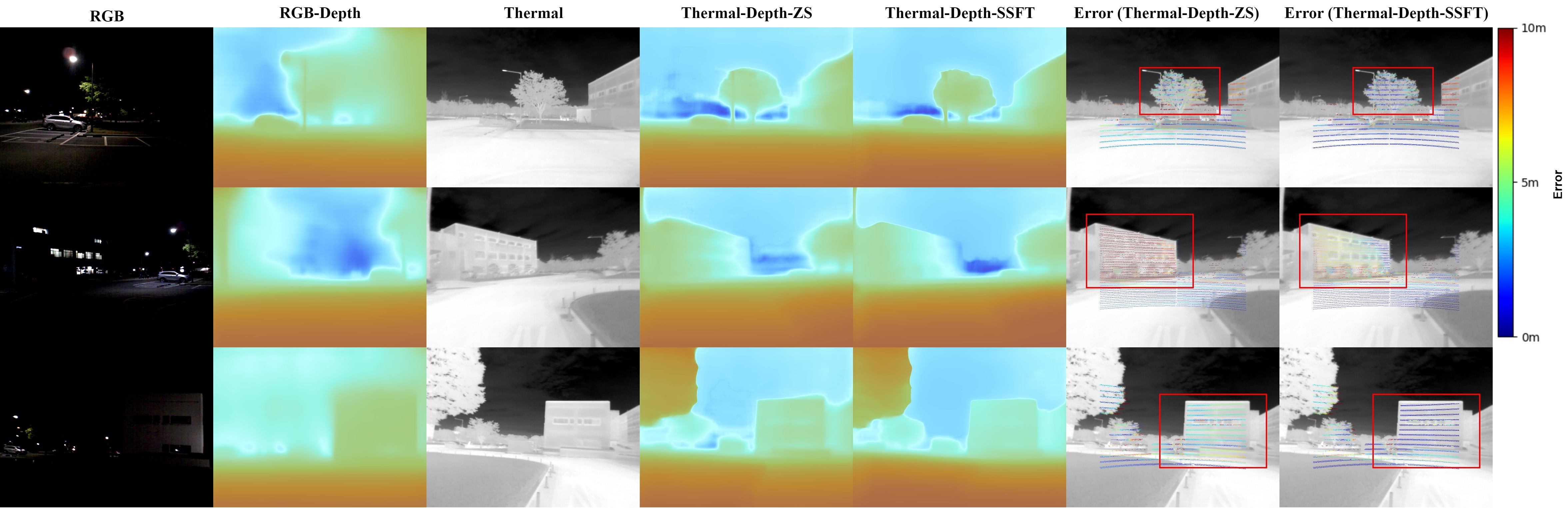}
  \caption{\textbf{Monocular Depth Estimation on ViViD++ dataset~\cite{lee2022vivid++}.} From left to right: the RGB image, our predicted RGB depth, the normalized thermal image, our predicted thermal depth with zero-shot (Thermal-Depth-ZS), our predicted thermal depth after self-supervised fine-tuning (Thermal-Depth-SSFT), the depth error of Thermal-Depth-ZS, and the depth error of Thermal-Depth-SSFT. The red boxes highlight areas with a significant decrease in error after self-supervised fine-tuning.}
  \label{fig:vivid}
\end{figure*}

\PAR{Zero-Shot Generalization on ViViD++.}
To evaluate zero-shot generalization, we assess various methods on the outdoor test split of the ViViD++ dataset. All networks are evaluated with weights trained on the MS$^2$ dataset. The results for unweighted metrics are presented in the top part of Table~\ref{tab:vividoutdoor}, with qualitative results shown in Fig.~\ref{fig:vivid}. We report the zero-shot performance of both our RGB MDE model and thermal MDE model, denoted `Ours-ZS'. Since the test sequences are collected at night, the thermal MDE model significantly outperforms the RGB MDE model, achieving RMSE values of 4.440 and 5.903, respectively. 

Additionally, we report the results of our method trained without RGB-to-thermal distillation, denoted as `Ours-NoDist-ZS'. It is evident that `Ours-ZS' with distillation demonstrates superior performance and better generalization compared to `Ours-NoDist-ZS', highlighting the effectiveness of our distillation approach.

\begin{table*}[tb!]
	\begin{center}
            \caption{\protect\label{tab:vividoutdoor}\protect\textbf{Generalization and self-supervised fintuning test on ViViD++ outdoor dataset~\protect\cite{lee2022vivid}.} `ZS' means `Zero Shot', and `SSFT' means `Self supervised finetuning.' Unweighted metrics are shown.
		}
		\renewcommand\arraystretch{0.99}
		\setlength\tabcolsep{8pt} 
		\small
		\begin{tabular}{c|c|c|cccc|ccc}
			\hline
			\multirow{2}{*}{\textbf{Split}} &
			\multirow{2}{*}{\textbf{Method}} &
			\multirow{2}{*}{\textbf{Modality}}&
			\multicolumn{4}{c|}{Error $\downarrow$}&
			\multicolumn{3}{c}{Accuracy $\uparrow$}\\
			\cline{4-10}
			&&&AbsRel&SqRel&RMSE&RMSElog&$\delta<1.25$&$\delta<1.25^{2}$&$\delta<1.25^{3}$\\
			\hline
			\hline
			\multirow{6}{*}{Test}&
			ZoeDepth~\cite{zoedepth2023}   & Ther & 0.174 & 1.163 & 5.633 & 0.227 & 0.685 & 0.944 & 0.989 \\
			& DepthAnyting~\cite{yang2024depth} & Ther & 0.166 & 0.937 & 4.929 & 0.204 & 0.719 & 0.973 & 0.994 \\
			& Ours-NoDist-ZS        & Ther & 0.154 & 0.798 & 4.540 & 0.187 & 0.758 & 0.983 & \textbf{0.997} \\
			& Ours-ZS               & RGB  & 0.198 & 2.037 & 5.903 & 0.241 & 0.710 & 0.940 & 0.978 \\
			& Ours-ZS               & Ther & 0.153 & 0.801 & 4.440 & 0.184 & 0.768 & 0.984 & \textbf{0.997} \\
			& Ours-SSFT             & Ther & \textbf{0.118} & \textbf{0.593} & \textbf{4.006} & \textbf{0.147} & \textbf{0.897} & \textbf{0.988} & \textbf{0.997} \\
			\hline
			\hline
			\multirow{2}{*}{Training}&
			\cellcolor{gray}Ours-ZS  & \cellcolor{gray}Ther & \cellcolor{gray}0.164 & \cellcolor{gray}1.042 & \cellcolor{gray}5.335 &  \cellcolor{gray}0.203 & \cellcolor{gray}0.718 & \cellcolor{gray}0.971 & \cellcolor{gray}0.996 \\
			&\cellcolor{gray}Ours-ZS      & \cellcolor{gray}RGB  & \cellcolor{gray}0.124 & \cellcolor{gray}0.641 & \cellcolor{gray}4.199 & \cellcolor{gray}0.153 & \cellcolor{gray}0.883 & \cellcolor{gray}0.988 & \cellcolor{gray}0.997 \\
			\hline
		\end{tabular}
		\vspace{-2em}
	\end{center}
\end{table*}
\PAR{Self-Supervised Finetuning on ViViD++.}
One of the key advantages of our framework is its ability to perform self-supervised fine-tuning of the thermal model using the RGB model, especially in new applications where ground-truth depth is not available. 
Before fine-tuning, we evaluated the zero-shot performance of our method on the outdoor training split (captured at daytime) of the ViViD++ dataset, with results shown in Table~\ref{tab:vividoutdoor}. The RGB model demonstrates good generalization, achieving an RMSE of 4.199, which is much lower than the thermal model's RMSE of 5.335. Therefore, the RGB model is able to teach the thermal model and improve its performance.

We fine-tuned our thermal MDE model using the RGB MDE model on the training split of the ViViD++ dataset through our proposed confidence-aware distillation. Notably, no ground-truth information was used during the self-supervised fine-tuning process. After fine-tuning, the performance of the thermal MDE model significantly improved, as denoted by `Ours-SSFT' in Table~\ref{tab:vividoutdoor}. Compared to the zero-shot thermal model `Ours-ZS', `Ours-SSFT' shows a substantial reduction in AbsRel from 0.153 to 0.118 (a 22.88\% decrease) and an improvement in threshold accuracy $(\delta <1.25)$ from 76.8\% to 89.7\%. The predicted depth from `Ours-ZS' and `Ours-SSFT' and their error maps are visualized in Fig.~\ref{fig:vivid}. These results show that our confidence-aware distillation method effectively enhances thermal MDE performance in new scenarios, which is crucial for deploying thermal MDE models in novel scenarios.

\vspace{-16pt}\subsection{Ablation Study}
We conducted an ablation study on the MS$^2$ dataset to evaluate the impact of various design choices in our framework. The following configurations were examined:
(1) No Distillation: This configuration trains the thermal and RGB MDE models together using their respective ground-truth depth maps, without any distillation step.
(2) No Confidence Network: In this setup, we perform distillation from the RGB model to the thermal model without incorporating the confidence network for weighting.
(3) No Multi-Modal Confidence: Here, we remove all input components related to the thermal branch when predicting the confidence of the RGB MDE, using only RGB-related metadata in the confidence network.

The results of these configurations are summarized in Table~\ref{tab:ablation}, where they are denoted as `No Dist.', `No Conf.', and `No Mul. Conf.', respectively. Our full method, which includes all design choices, achieves the best performance.
Interestingly, the results reveal that distillation without the predicted confidence (`No Conf.') is detrimental to thermal MDE performance, even showing significantly worse results compared to the setup without any distillation. This underscores the importance of confidence-aware distillation.
\begin{table}[tb!]
	\begin{center}
            \caption{\label{tab:ablation} \textbf{The ablation study for our design choices.} The weighted metrics are shown.} 
		\renewcommand\arraystretch{0.9}
		\setlength\tabcolsep{5.0pt} 
		\small
		\begin{tabular}{c|cc|c}
			
			\hline
			Setup
			&RMSE$\downarrow$
			&SqRel $\downarrow$
			&$\delta<1.25\uparrow$\\
			\hline
			\hline
			Ours & 4.330 & 0.619 & 0.925 \\ 
			No Dist. & 4.359 & 0.626 & 0.923 \\ 
			No Conf.& 4.469 & 0.664 & 0.919 \\ 
			No Mul. Conf.& 4.345 & 0.625 & 0.924\\ 
			
			\hline

\end{tabular}
\vspace{-0.5em}

\end{center}
\vspace{-20pt}
\end{table}

\vspace{-10pt}\subsection{Validation on Real Robots}
\begin{figure}\vspace{-15pt}
    \centering
    \begin{subfigure}[c]{.98\columnwidth}
        \includegraphics[width=\textwidth]{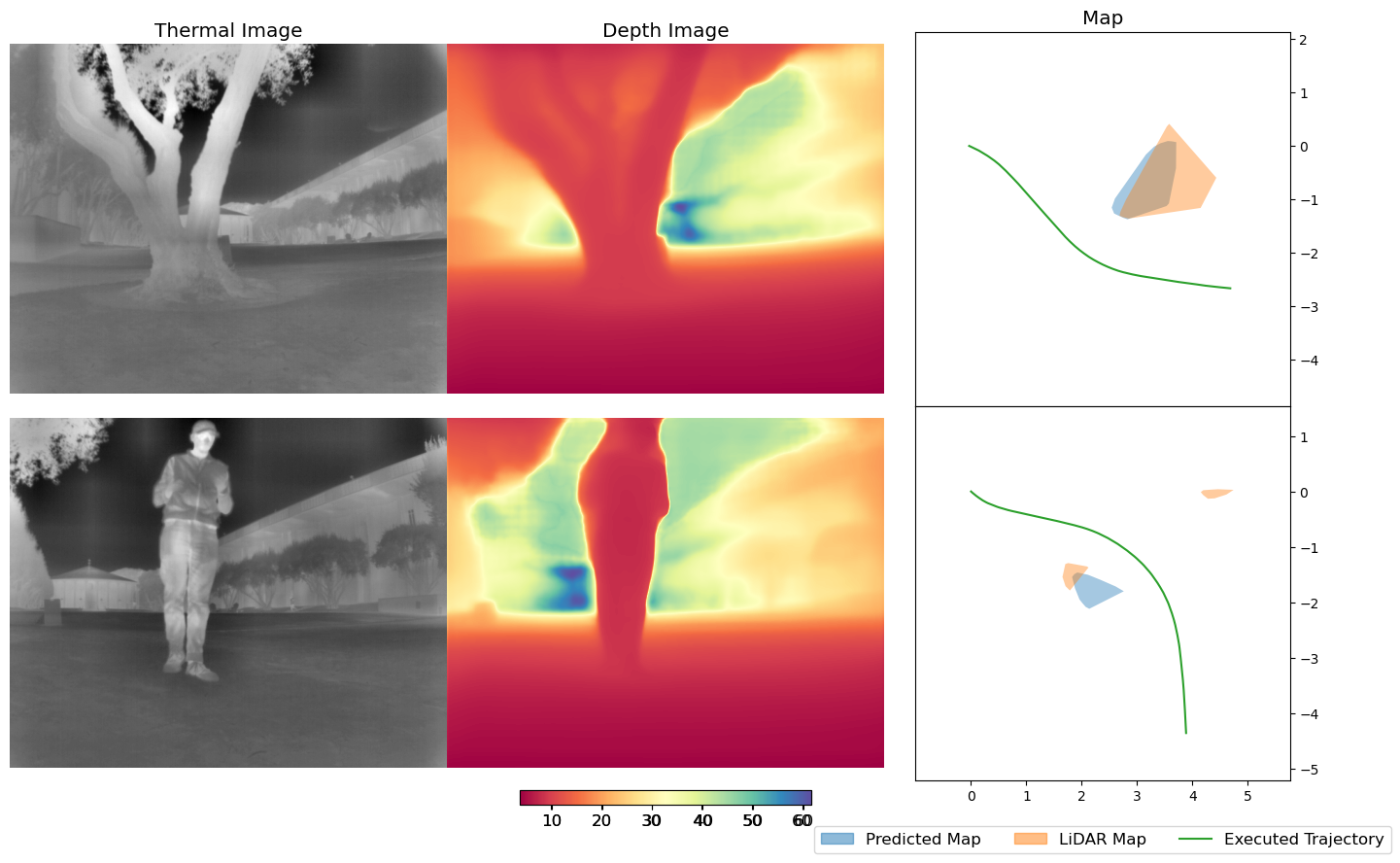}
    \end{subfigure}
    \caption{\textbf{Zero-shot deployment of MonoTher-Depth} onto a robotic hardware platform, a Traxxas-Maxx car equipped with a FLIR ADK thermal camera, Velodyne Puck LiDAR, and Simply NUC Ruby for computation. Two example demonstrations, in which the robot is tasked with avoiding an obstacle (a tree in the top series, a person in the bottom series). The resulting polygonal obstacle map, along with a map derived from ground-truth LiDAR data, is shown in the far right.}
    \label{fig:real-world}\vspace{-15pt}
    \end{figure}

We have conducted a series of qualitative zero-shot experiments at night on robotic hardware in a real-world setting, as shown in Fig.~\ref{fig:real-world}, where MonoTher-Depth is deployed in a zero-shot manner. The predicted depth image is projected into a point cloud, which is then processed with uniform downsampling and radial outlier rejection. Ground points are removed, and the resulting point cloud is flattened into a 2D plane. Finally, the points are clustered using the density-based spatial clustering (DBSCAN) algorithm~\cite{ester1996density} and converted into a map of convex polygons via the alphashape method \cite{akkiraju1995alphashape}.
The resulting 2D polygonal obstacle maps derived from thermal depths are shown on the far right of Fig.~\ref{fig:real-world} and are very close to the 2D polygonal obstacle maps obtained from 3D LiDAR point clouds. These 2D polygonal maps from thermal depths are then used for obstacle avoidance along a predefined path to a waypoint, successfully achieving collision-free navigation.
These experiments demonstrate the effectiveness of our proposed MonoTher-Depth in out-of-distribution settings for real-world robotic applications.

\vspace{-10pt}\subsection{Discussion and Limitations}

Our confidence-aware distillation method demonstrates impressive performance across scenarios with and without labeled depth supervision.
%
%
However, there are inherent limitations that MDE models trained on outdoor datasets struggle with indoor scenarios due to significant domain gaps. This outdoor-to-indoor generalization issue is also shown in~\cite{yang2024depth} and \cite{zoedepth2023}.
Similarly, our models trained on the MS$^2$ outdoor dataset exhibit poor zero-shot performance on the ViViD++ indoor scenarios~\cite{lee2022vivid}. The self-supervised fine-tuning using RGB-to-thermal distillation does not substantially improve performance in such cases, primarily due to the already poor predictions in indoor settings.

	\vspace{-12pt}\section{Conclusion}
\label{sec:conclusion}
We present MonoTher-Depth, a thermal monocular depth estimation (MDE) method that incorporates knowledge distilled from large, foundational RGB MDE models. To prevent the thermal MDE model from being adversely affected by the RGB MDE model in challenging scenarios, we introduce a novel confidence-aware distillation method. This approach adaptively adjusts the distillation strength based on the predicted confidence, utilizing both thermal and RGB information. By incorporating confidence-aware distillation, our thermal model achieves significant improvements in depth estimation accuracy, particularly in new scenarios where ground-truth depth supervision is unavailable.
	\vspace{-12pt}
	\bibliographystyle{IEEEtran}
	\bibliography{main}

\end{document}